\documentclass[journal]{IEEEtran}
\usepackage{hyperref}
\usepackage{cite}
\usepackage{amsmath,amssymb,amsfonts}
\usepackage{algorithmic}
\usepackage{graphicx}
\usepackage{textcomp}
\usepackage{xcolor}
\usepackage{booktabs}   
\usepackage{multirow}   
\usepackage{array}
\DeclareMathOperator*{\argmax}{arg\,max} 

\def\BibTeX{{\rm B\kern-.05em{\sc i\kern-.025em b}\kern-.08em
    T\kern-.1667em\lower.7ex\hbox{E}\kern-.125emX}}

\begin{document}

\title{Geospatial-Reasoning-Driven Vocabulary-Agnostic Remote Sensing Semantic Segmentation}



\author{Chufeng Zhou, Jian Wang, Xinyuan Liu, Jie Zheng, and Xiaokang Zhang,~\IEEEmembership{Senior Member,~IEEE} %
\thanks{This work was supported in part by the National Natural Science Foundation of China under Grant 42371374. (\textit{Corresponding author: Xiaokang Zhang}).}
\thanks{Chufeng Zhou and Jian Wang are with the School of Electronic Information, Wuhan University of Science and Technology, Wuhan 430081, China (e-mail:  zhouchufeng@wust.edu.cn; wangjian@wust.edu.cn).}
\thanks{Xinyuan Liu is with the State Key Laboratory of Networking and Switching Technology, Beijing University of Posts and Telecommunications, Beijing, China (e-mail: xinyuanliu@bupt.edu.cn).}
\thanks{Jie Zheng is with Oriental Space Port Research Institute, Yantai 265100, China (e-mail: zhengjie@ospri.org.cn).}
\thanks{Xiaokang Zhang is with the School of Artificial Intelligence, Wuhan University, Wuhan 430072, China (e-mail: zhangxiaokang@whu.edu.cn).}
}
\maketitle

\begin{abstract}
Open-vocabulary semantic segmentation has become an important direction in remote sensing, as it enables recognition beyond predefined land-cover categories. However, existing methods mainly depend on passive visual-text matching and often struggle with semantic ambiguity in geographically complex scenes, especially when different classes exhibit similar spectral or structural patterns. To address this issue, we propose a Geospatial Reasoning Chain-of-Thought (GR-CoT) framework for remote sensing open-vocabulary semantic segmentation. GR-CoT consists of an offline knowledge distillation stream and an online instance reasoning stream. The former constructs category interpretation standards for confusing classes, while the latter performs macro-scenario anchoring, visual feature decoupling, and knowledge-driven decision synthesis to generate an image-adaptive vocabulary for downstream segmentation. Experiments on the LoveDA and GID5 benchmarks indicate that the proposed framework improves overall segmentation performance and yields more semantically coherent predictions in complex scenes.
\end{abstract}


\begin{figure*}[htbp]
  \centering
  \includegraphics[width=1\linewidth]{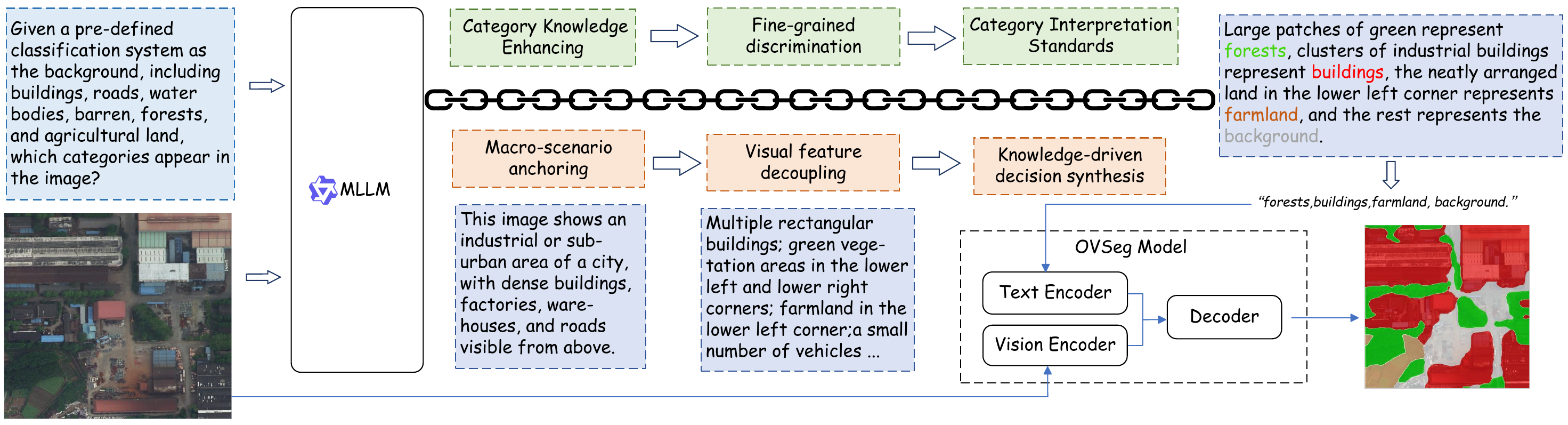}
  \caption{The proposed framework of geospatial reasoning chain-of-thought (GR-CoT) for remote sensing semantic segmentation. The architecture consists of two collaborative streams: an offline knowledge distillation stream (top) and an online instance reasoning stream (bottom). The offline stream establishes category interpretation standards via fine-grained discrimination to resolve semantic ambiguity. The online stream sequentially performs macro-scenario anchoring and visual feature decoupling, which are integrated in the knowledge-driven decision synthesis stage to generate an image-adaptive vocabulary. This refined vocabulary is then utilized by an open-vocabulary segmentation model to produce the final semantic mapping results.}
  \label{fig:framework}
\end{figure*}

\section{Introduction}

The rapid development of remote sensing technology has led to an explosion of high-resolution geospatial data, necessitating advanced semantic segmentation techniques for applications such as urban planning and environmental monitoring. Recently, open-vocabulary semantic segmentation has emerged as a transformative paradigm, enabling the identification of land-cover categories beyond fixed, predefined sets. Vision-language models such as CLIP \cite{radford2021learning} have established a fundamental bridge between visual features and textual semantics. Early methods such as LSeg \cite{li2022languagedriven} and OpenSeg \cite{ghiasi2022scaling} extended vision-language alignment to dense prediction, laying the foundation for open-vocabulary segmentation. Building upon this foundation, CAT-Seg \cite{cho2024cat} provides a robust framework that achieves effective open-vocabulary semantic segmentation through cost aggregation. In the specific context of remote sensing, several specialized methods \cite{cao2025open, li2026exploring, li2025segearth, liu2025large} have been proposed to address domain-specific challenges, offering effective solutions in terms of rotation invariance, scale variations, category understanding, and parameter efficiency. For instance, TPOV-Seg \cite{zhang2025tpov} introduces a text-guided category aggregator to significantly strengthen the model's recognition and generalization capabilities for unseen categories within complex geospatial environments.

Parallel to these advancements, Chain-of-Thought (CoT) prompting has emerged as a transformative technique to elicit structured, multi-step reasoning in large language models \cite{wei2022chain, kojima2022large}. By decomposing a complex problem into a sequence of intermediate logical steps, CoT mimics human-like cognitive processes to resolve task-specific ambiguities. This reasoning paradigm has been successfully extended to multimodal contexts \cite{zhang2023multimodal}, enabling MLLMs to synergize visual perception with high-level semantic deduction. 

However, a fundamental limitation persists in current open-vocabulary frameworks: they predominantly rely on a passive visual-semantic matching paradigm. These models identify objects primarily based on morphological and spectral similarities—essentially a "looks-like" approach—without a deep understanding of the geospatial context or functional attributes of the land cover. In complex remote sensing scenes, this leads to significant semantic ambiguity. Without incorporating higher-level geographical logic or macro-scene priors, passive matching models are prone to frequent misclassifications.

In this paper, we investigate the open-vocabulary semantic segmentation problem by proposing a geospatial reasoning chain-of-thought (GR-CoT) framework. Our approach shifts the paradigm from passive matching to active geographical reasoning by integrating the logical reasoning capabilities of MLLMs. The main contributions of this work are summarized as follows. First, we propose a reasoning-driven mechanism to resolve semantic ambiguity in remote sensing open-vocabulary segmentation by transforming passive visual-semantic matching into an active, logic-driven process. Second, we develop a collaborative dual-stream architecture that synergizes offline knowledge distillation for category standards with online instance reasoning for dynamic scene understanding. The framework executes a sequential reasoning process—including macro-scenario anchoring and visual feature decoupling—to generate an image-adaptive vocabulary, ensuring the downstream segmentation is guided by correct geographical logic. Finally, experiments on the LoveDA and GID5 benchmarks show that the proposed GR-CoT improves overall performance and yields semantically coherent qualitative results in diverse remote sensing scenes.


\section{Methodology}

The proposed geospatial reasoning chain-of-thought (GR-CoT) framework is designed to transform the traditional passive visual-semantic matching into an active, logic-driven process. Given an input remote sensing image $I$, the objective is to generate a pixel-level semantic segmentation map $M$ by leveraging an image-adaptive vocabulary $V_{adaptive}$ derived from a structured reasoning chain. As shown in Fig.~\ref{fig:framework}, the framework comprises an offline knowledge distillation stream and an online instance reasoning stream, which collaborate to resolve semantic ambiguities in complex geographical environments. In our implementation, the multimodal large language model (MLLM) is instantiated as Qwen-VL-Max, a member of the Qwen-VL family \cite{bai2025qwen3}. It is used in both the offline knowledge distillation stream and the online instance reasoning stream to produce category-aware textual reasoning outputs, which are then converted into an image-adaptive candidate vocabulary for downstream open-vocabulary segmentation.

\subsection{Offline Knowledge Distillation Stream}
The offline knowledge distillation stream distills expert priors into a Category interpretation standards $S$ to provide a cognitive foundation for land-cover interpretation. Initially, the MLLMs perform category knowledge enhancement for each class in the global category pool $C = \{c_{1}, c_{2}, \dots, c_{n}\}$. In this stage, the MLLMs is prompted to describe detailed geographical attributes, including geometric shapes, boundary contours, and typical sub-classes, to establish a rich descriptive foundation.

To further mitigate semantic conflicts between morphologically similar classes, a fine-grained discrimination process is executed where the MLLMs defines rigorous relationships for categories prone to semantic ambiguity. For instance, the reasoning chain clarifies that structures such as steel-framed greenhouses and plastic mulch should be categorized as agricultural land rather than industrial buildings, and that isolated bare land with messy surface textures belongs to the barren category rather than active farmland. In the final stage, the enhanced geographical knowledge and the results of fine-grained inter-class discrimination are provided as input to the MLLMs to synthesize the Category Interpretation Standards $D_{i}$ for each class:
\begin{equation}
S = \{ (c_{i}, D_{i}) \mid c_{i} \in C, D_{i} = \text{MLLMs}(c_{i}, \text{priors}) \},
\end{equation}
where $D_{i}$ encapsulates a multi-dimensional standard covering morphology, spectral-spatial attributes, and spatial exclusivity, thereby forming the finalized category interpretation standards.

\subsection{Online instance reasoning stream}
The online stream executes a sequential chain-of-thought to bridge the gap between low-level visual facts and high-level geographical logic through three distinctive stages.

The first stage is macro-scenario anchoring, which identifies the global context $G$ of the image $I$:
\begin{equation}
G = f_{anchor}(I) \in \{ \text{urban, rural, industrial, \dots} \}.
\end{equation}
This context $G$ establishes a geographical prior that constrains the potential category space. Subsequently, visual feature decoupling is performed to decompose the scene into a set of discrete visual attributes $A = \{a_1, a_2, \dots, a_m\}$:
\begin{equation}
A = f_{decouple}(I, G),
\end{equation}
where each attribute $a_{j}$ describes objective characteristics such as geometric textures, spectral reflectance, and fine-grained categories.

The final stage is knowledge-driven decision synthesis, which integrates the image-specific context $G$, visual facts $A$, and Category interpretation standards $S$ to form the image-adaptive vocabulary $V_{adaptive}$:
\begin{equation}
V_{adaptive} = \{ c_i \mid \text{verify}(c_i, G, A, S) = \text{true} \}.
\end{equation}
The verification function ensures that the selected categories are logically consistent with the geographical environment, such as correctly identifying agricultural structures in rural contexts rather than industrial buildings.

\begin{table*}[t]
\centering
\caption{Structured prompt design of the proposed GR-CoT framework.}
\label{tab:prompt_design}
\renewcommand{\arraystretch}{1.25}
\setlength{\tabcolsep}{5pt}
\footnotesize
\begin{tabular}{@{}
>{\raggedright\arraybackslash}p{0.17\textwidth}
>{\raggedright\arraybackslash}p{0.22\textwidth}
>{\raggedright\arraybackslash}p{0.55\textwidth}
@{}}
\toprule
\textbf{Prompt Stage} & \textbf{Input} & \textbf{Required Output} \\
\midrule

Offline category interpretation
&
Global category set
&
Category definition, visual features, geographical context, inclusion/exclusion criteria, and confusing categories with decision rules. \\

\midrule

Online image reasoning
&
Input remote sensing image and global category set
&
Macro-scenario type, supporting evidence, likely/uncertain categories, geometric features, texture features, spectral appearance, spatial relationships, and ambiguous observations. \\

\midrule

Knowledge-driven vocabulary synthesis
&
Global category set, category interpretation standards, and online image reasoning result
&
Selected categories, excluded confusing categories, uncertain categories, and the final image-adaptive vocabulary. \\

\bottomrule
\end{tabular}
\end{table*}

\subsection{Knowledge-driven open-vocabulary segmentation}
The candidate categories produced by the multimodal large language model through the reasoning chain are used to construct an adaptive vocabulary set, denoted as $V_{adaptive}$. This set is then fed into the downstream open-vocabulary segmentation model as the candidate label space. Given $V_{adaptive}$, the downstream segmentation stage follows the standard pixel-to-text alignment paradigm:
\begin{equation}
M(x, y) = \argmax_{c_j \in V_{adaptive}} \langle \mathbf{F}_v(x, y), \mathbf{E}_t(c_j) \rangle,
\end{equation}
where $\mathbf{F}_v(x, y)$ denotes the visual feature at pixel $(x, y)$ extracted by the vision encoder, and $\mathbf{E}_t(c_j)$ denotes the text embedding of category $c_j$ generated by the text encoder. In this way, the reasoning chain does not directly perform pixel-level segmentation; instead, it narrows the candidate category space for the downstream OVSeg model. In practice, the category names predicted by Qwen-VL-Max are mapped to the label set supported by the downstream OVSeg model and then used as candidate categories for segmentation. By restricting inference to the reasoning-generated vocabulary, the framework helps reduce cross-category confusion and improves the semantic consistency of pixel-level predictions in geographically complex scenes.

\subsection{Prompt Design for Geospatial Reasoning}

To make the proposed reasoning process explicit and reproducible, we design three structured prompts corresponding to the three reasoning stages of GR-CoT, as summarized in Table~\ref{tab:prompt_design}. The first prompt is used in the offline knowledge distillation stream, where the MLLM is required to construct category interpretation standards for each category in the global category set. Specifically, the model is instructed to describe the category definition, visual features, geographical context, inclusion criteria, exclusion criteria, and confusing categories with fine-grained decision rules. This prompt converts general land-cover names into interpretable geospatial knowledge that can be reused across images.

The second prompt is used in the online instance reasoning stream. Given an input remote sensing image, the MLLM first performs macro-scenario anchoring to identify the dominant geographical context, such as urban, rural, agricultural, forest-dominated, water-dominated, or mixed scenes. It then conducts visual feature decoupling by describing geometric structures, texture patterns, color and spectral appearances, spatial relationships, and potential fine-grained objects. Importantly, this stage does not directly output the final category vocabulary, but only records observable visual evidence and uncertain regions.

The third prompt is designed for knowledge-driven decision synthesis. It takes the global category set, the offline category interpretation standards, and the online image reasoning result as inputs, and determines which categories are truly present in the image. The MLLM is required to select only categories supported by visual evidence and consistent with the macro-scenario prior, while excluding visually similar but geographically inconsistent categories. The output is a JSON-formatted image-adaptive vocabulary, which is then mapped to the candidate label set of the downstream open-vocabulary segmentation model.

\section{Experiments}

We evaluate the proposed GR-CoT framework on the LoveDA \cite{loveda} and GID5 \cite{tong2020land} datasets. The experiments include qualitative visualization, quantitative comparison, and ablation analysis to verify the effectiveness of the proposed reasoning-guided vocabulary construction strategy.

\subsection{Datasets}

\subsubsection{LoveDA}
LoveDA \cite{loveda} is a land-cover semantic segmentation dataset designed for unsupervised domain adaptation and cross-domain semantic segmentation. It contains urban and rural subsets with high-resolution RGB images of size $1024\times1024$ pixels, covering seven land-cover categories: Background, Building, Road, Water, Barren, Forest, and Agriculture. In this work, we use the official test split for evaluation, including 677 urban images.

\subsubsection{GID}
The Gaofen Image Dataset (GID) \cite{tong2020land} contains high-resolution imagery captured by the Gaofen-2 satellite and supports large-scale remote sensing land-cover applications. It includes two subsets: GID-5 for broad land-cover classification and GID-15 for more fine-grained annotations. In this work, we evaluate our method on GID5, where 420 cropped image patches of size $1024\times1024$ pixels are used for testing.

\begin{table*}[t]
\centering
\caption{Quantitative Comparisons Evaluated on LoveDA(\%)}
\label{tab:loveda_comparison}
\renewcommand{\arraystretch}{1.2}
\resizebox{1.0\textwidth}{!}{
\begin{tabular}{l | cc cc cc cc cc cc cc | cc}
\toprule
\multirow{2}{*}{Methods} & 
\multicolumn{2}{c}{Agricultural} & 
\multicolumn{2}{c}{Background} & 
\multicolumn{2}{c}{Barren} & 
\multicolumn{2}{c}{Building} & 
\multicolumn{2}{c}{Forest} & 
\multicolumn{2}{c}{Road} & 
\multicolumn{2}{c}{Water} & 
\multicolumn{2}{c}{Overall} \\
\cmidrule(lr){2-3} \cmidrule(lr){4-5} \cmidrule(lr){6-7} \cmidrule(lr){8-9} \cmidrule(lr){10-11} \cmidrule(lr){12-13} \cmidrule(lr){14-15} \cmidrule(lr){16-17}
 & IoU & Acc & IoU & Acc & IoU & Acc & IoU & Acc & IoU & Acc & IoU & Acc & IoU & Acc & mIoU & OA \\
\midrule
CAT-Seg \cite{cho2024cat} & 46.54 & 54.31 & 0.19 & 0.19 & 16.02 & 16.64 & 39.12 & 92.20 & 38.40 & \textbf{85.90} & 37.14 & 82.27 & 62.19 & 81.42 & 34.23 & 51.75 \\
RSKT-Seg \cite{li2026exploring} & 57.91 & 73.53 & 2.65 & 2.78 & 9.71 & 9.90 & \textbf{46.75} & 93.29 & 45.33 & 80.80 & 36.89 & 87.00 & 61.95 & 88.76 & 36.82 & 56.85 \\
Ours & \textbf{61.19} & \textbf{77.44} & \textbf{10.57} & \textbf{11.20} & \textbf{16.81} & \textbf{17.45} & 46.35 & \textbf{95.79} & \textbf{51.53} & 79.35 & \textbf{44.14} & \textbf{85.00} & \textbf{67.14} & \textbf{90.42} & \textbf{41.39} & \textbf{59.93} \\
\bottomrule
\end{tabular}
}
\end{table*}

\begin{table*}[t]
\centering
\caption{Quantitative Comparisons Evaluated on GID5 (\%)}
\label{tab:gid5_comparison}
\renewcommand{\arraystretch}{1.2}
\resizebox{1.0\textwidth}{!}{
\begin{tabular}{l | cc cc cc cc cc cc | cc}
\toprule
\multirow{2}{*}{Methods} & 
\multicolumn{2}{c}{Background} & 
\multicolumn{2}{c}{Built-up} & 
\multicolumn{2}{c}{Farmland} & 
\multicolumn{2}{c}{Forest} & 
\multicolumn{2}{c}{Meadow} & 
\multicolumn{2}{c}{Water} & 
\multicolumn{2}{c}{Overall} \\
\cmidrule(lr){2-3} \cmidrule(lr){4-5} \cmidrule(lr){6-7} \cmidrule(lr){8-9} \cmidrule(lr){10-11} \cmidrule(lr){12-13} \cmidrule(lr){14-15}
 & IoU & Acc & IoU & Acc & IoU & Acc & IoU & Acc & IoU & Acc & IoU & Acc & mIoU & OA \\
\midrule
CAT-Seg \cite{cho2024cat} & 10.77 & 11.47 & 51.14 & 74.73 & 63.82 & 88.19 & 58.84 & 84.11 & 14.38 & 57.07 & 40.43 & 91.78 & 40.23 & 59.12 \\
RSKT-Seg \cite{li2026exploring} & \textbf{46.96} & \textbf{89.33} & 29.46 & 31.38 & 55.52 & 56.98 & \textbf{67.91} & 83.65 & 4.05 & 4.06 & 45.27 & 55.47 & 42.20 & 61.40 \\
Ours & 19.77 & 21.83 & \textbf{52.60} & \textbf{75.91} & \textbf{65.32} & \textbf{91.45} & 60.11 & \textbf{86.23} & \textbf{26.60} & \textbf{64.18} & \textbf{47.68} & \textbf{92.06} & \textbf{45.34} & \textbf{63.34} \\
\bottomrule
\end{tabular}
}

\end{table*}

\subsection{Evaluation Metrics}

We adopt mean Intersection over Union (mIoU) and overall accuracy (OA) as the primary metrics for semantic segmentation evaluation. Given $C$ semantic categories, mIoU is defined as:
\begin{equation}
\mathrm{mIoU}=\frac{1}{C}\sum_{i=1}^{C}\frac{TP_i}{TP_i+FP_i+FN_i},
\end{equation}
where $TP_i$, $FP_i$, and $FN_i$ denote the true positives, false positives, and false negatives of class $i$, respectively.

OA measures the proportion of correctly classified pixels among all pixels:
\begin{equation}
\mathrm{OA}=\frac{\sum_{i=1}^{C}n_{ii}}{\sum_{i=1}^{C}\sum_{j=1}^{C}n_{ij}},
\end{equation}
where $n_{ij}$ denotes the number of pixels belonging to class $i$ but predicted as class $j$. In the ablation study, we further report Category Accuracy (Cat. Acc.) to evaluate whether the correct land-cover categories present in an image can be identified by the dynamic vocabulary.

\begin{figure}[htbp] 
  \centering
  \includegraphics[width=1.0\columnwidth]{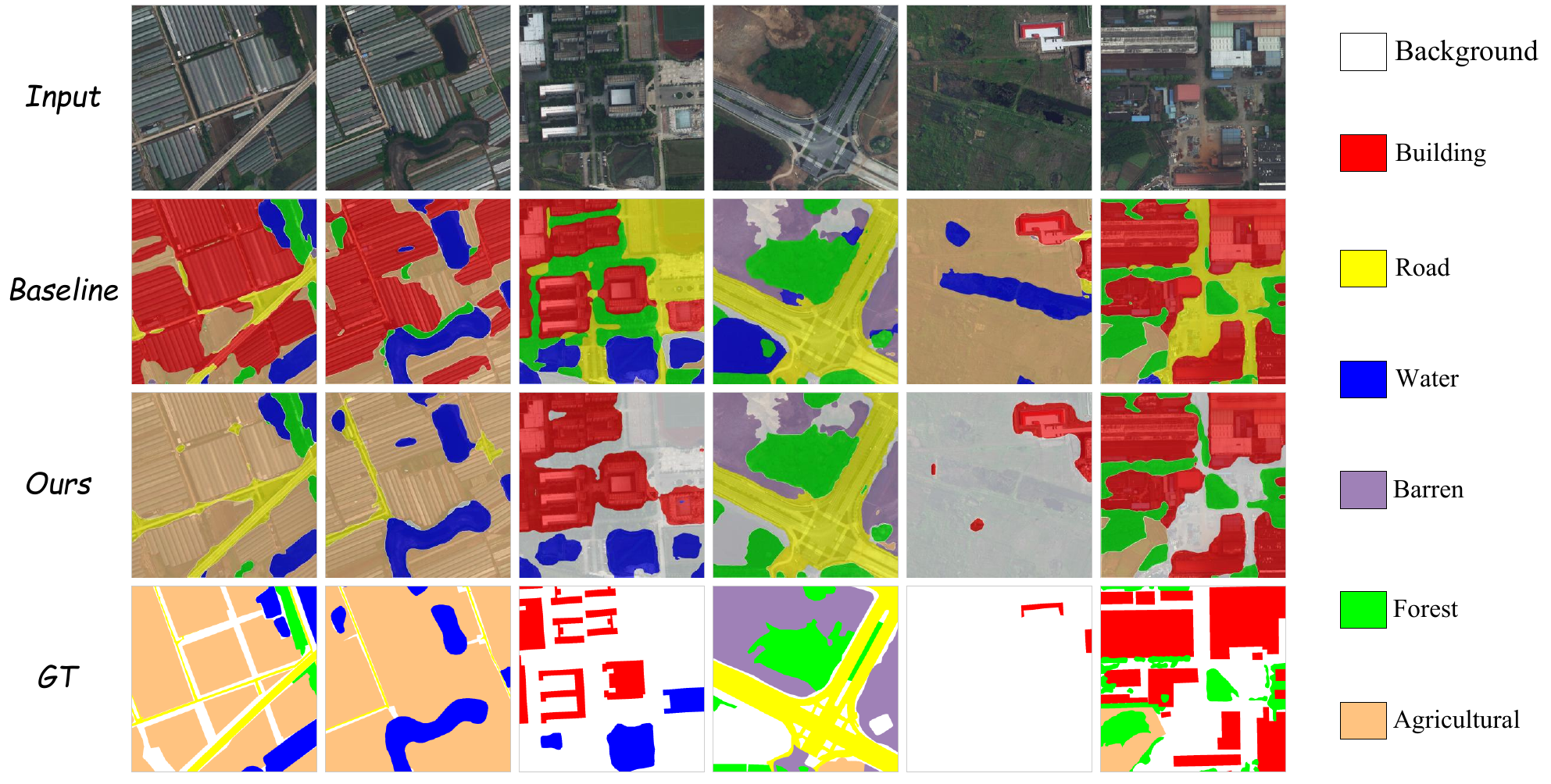} 
  \caption{Visualized results on the LoveDA dataset.}
  \label{fig:loveda_vis}
\end{figure}

\begin{figure}[htbp] 
  \centering
  \includegraphics[width=1.0\columnwidth]{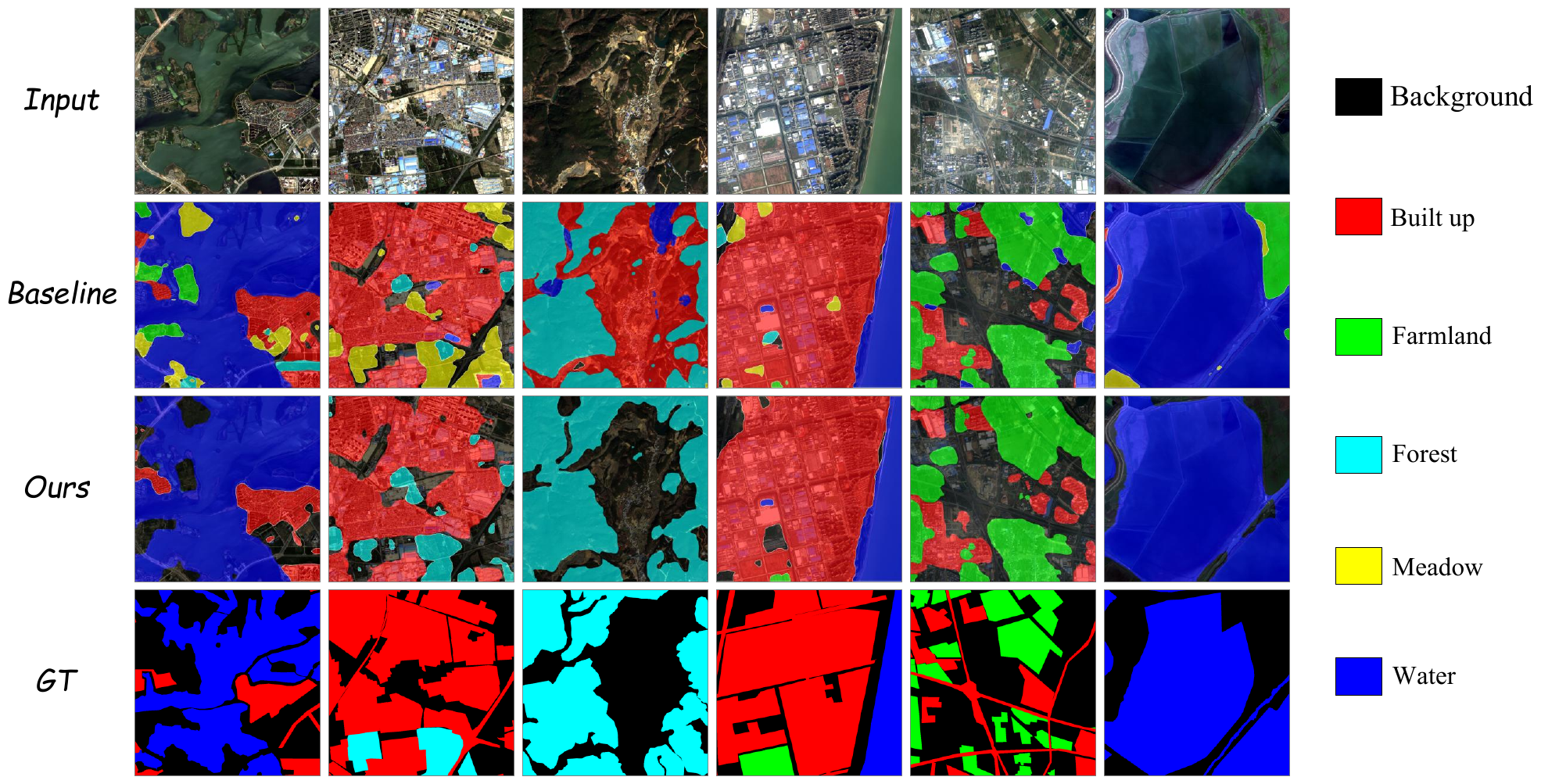} 
  \caption{Visualized results on the GID5 dataset.}
  \label{fig:gid5_vis}
\end{figure}

\subsection{Qualitative Evaluation}

As illustrated in Fig.~\ref{fig:loveda_vis} and Fig.~\ref{fig:gid5_vis}, the proposed GR-CoT framework produces more semantically coherent segmentation results than the RSKT-Seg baseline. In LoveDA scenes, the baseline tends to confuse regular agricultural structures, fragmented rural regions, and shadowed areas with buildings or water bodies. By introducing macro-scenario anchoring and category interpretation standards, GR-CoT better distinguishes agricultural land from building-like objects and suppresses visually plausible but geographically inconsistent predictions.

Similar improvements can be observed on the GID5 dataset. In forest- and built-up-area-dominated scenes, the baseline produces false positives in built-up or background regions, while GR-CoT generates cleaner and more consistent masks for meadow, forest, and water. These qualitative results indicate that reasoning-guided vocabulary generation helps reduce semantic ambiguity in complex remote sensing scenes.

\subsection{Quantitative Evaluation}

The quantitative evaluation results on the LoveDA and GID5 benchmarks are summarized in Table~\ref{tab:loveda_comparison} and Table~\ref{tab:gid5_comparison}, respectively. We compare the proposed GR-CoT framework with representative open-vocabulary segmentation methods, including CAT-Seg \cite{cho2024cat} and RSKT-Seg \cite{li2026exploring}. The comparison is conducted using class-wise IoU and accuracy, as well as overall mIoU and OA.

As shown in Table~\ref{tab:loveda_comparison}, GR-CoT achieves the best overall performance on LoveDA, with 41.39\% mIoU and 59.93\% OA, improving RSKT-Seg by 4.57\% and 3.08\%, respectively. At the category level, GR-CoT obtains the highest IoU on most classes, including agricultural land, background, barren land, forest, road, and water. This demonstrates its advantage in reducing semantic confusion, especially for fragmented or visually ambiguous regions. Although RSKT-Seg achieves a slightly higher IoU on the building class, GR-CoT obtains the best building accuracy, indicating that it remains competitive in recognizing man-made structures.

Table~\ref{tab:gid5_comparison} reports the results on GID5. GR-CoT achieves the best overall performance with 45.34\% mIoU and 63.34\% OA. It obtains clear improvements on built-up, farmland, meadow, and water categories, showing that the proposed reasoning-guided vocabulary is effective for distinguishing land-cover classes with diverse spatial patterns and visual appearances. Overall, the consistent gains on both datasets verify the effectiveness of GR-CoT in improving category selection and open-vocabulary segmentation performance.

\subsection{Ablation Study}

To verify the contribution of each component, we conduct ablation experiments as summarized in Table~\ref{tab:ablation}. The three configurations correspond to the plain OVSeg baseline, OVSeg with MLLM-enhanced category knowledge, and the full GR-CoT framework, respectively. The plain OVSeg baseline obtains only 11.19\% Cat. Acc., indicating that passive visual-text matching suffers from severe semantic ambiguity in remote sensing scenes.

By introducing MLLM-derived category knowledge, Cat. Acc. increases to 37.33\%, showing that enhanced class descriptors help the model better identify plausible land-cover categories. However, this setting does not achieve the best OA, suggesting that category-level knowledge alone may still introduce noisy or over-inclusive vocabularies. In contrast, the full GR-CoT framework further incorporates macro-scenario anchoring and visual feature decoupling, achieving the best Cat. Acc., mIoU, and OA. These results demonstrate that image-specific geospatial reasoning is essential for generating reliable adaptive vocabularies and improving final segmentation quality.

\begin{table}[h]
\centering
\caption{Ablation study of the GR-CoT framework (\%). $\checkmark$ indicates the component is active. ``Cat. Acc.'' denotes the accuracy of identifying the correct categories present in the image.}
\label{tab:ablation}
\setlength{\tabcolsep}{1pt}
\resizebox{0.9\columnwidth}{!}{
\begin{tabular}{>{\centering\arraybackslash}p{1.8cm} >{\centering\arraybackslash}p{1.2cm} >{\centering\arraybackslash}p{1.2cm} >{\centering\arraybackslash}p{1.8cm} >{\centering\arraybackslash}p{1.4cm} >{\centering\arraybackslash}p{1.4cm}}
\hline
OVSeg Model & MLLMs & GR-CoT & Cat. Acc. & mIoU & OA \\
\hline
$\checkmark$ &              &                      & 11.19    & 42.20 & 61.40 \\
$\checkmark$ & $\checkmark$ &                      & 37.33    & 43.31 & 61.32 \\
$\checkmark$ & $\checkmark$ & $\checkmark$         & \textbf{45.59} & \textbf{45.34} & \textbf{63.34} \\
\hline
\end{tabular}
}
\end{table}

\section{Conclusion}
In this paper, we propose a framework named Geospatial Reasoning Chain-of-Thought (GR-CoT), designed to address the persistent semantic ambiguity and the challenge of distinguishing spectrally similar land-cover types in open-vocabulary remote sensing semantic segmentation. By shifting the paradigm from passive visual-semantic matching to active geographical reasoning, 
our approach leverages the logical reasoning capabilities of multimodal large language models to introduce expert-inspired geospatial reasoning into the segmentation process. The dual-stream architecture—comprising an offline knowledge distillation stream and an online instance reasoning stream—enables the generation of image-adaptive vocabularies grounded in macro-scenario context and fine-grained visual facts, which subsequently guide the model toward accurate segmentation. Experimental results on the LoveDA and GID5 benchmarks show that GR-CoT improves overall performance and yields more semantically consistent results, particularly when encountering spectrally similar and fine-grained land-cover types. This work highlights the critical role of geospatial logic in advancing robust scene understanding for open-vocabulary remote sensing applications.

\bibliographystyle{IEEEtran}
\bibliography{refs}
\end{document}